# Measuring the Hardness of Stochastic Sampling on Bayesian Networks with Deterministic Causalities: the $k$-Test

**Haohai Yu** and **Robert van Engelen**
Department of Computer Science
Florida State University
Tallahassee, FL 32306-4530

## Abstract

Approximate Bayesian inference is **NP**-hard. Dagum and Luby defined the *Local Variance Bound* (LVB) to measure the approximation hardness of Bayesian inference on Bayesian networks, assuming the networks model strictly positive joint probability distributions, i.e. zero probabilities are not permitted. This paper introduces the $k$-test to measure the approximation hardness of inference on Bayesian networks with deterministic causalities in the probability distribution, i.e. when zero conditional probabilities are permitted. Approximation by stochastic sampling is a widely-used inference method that is known to suffer from inefficiencies due to sample rejection. The $k$-test predicts when rejection rates of stochastic sampling a Bayesian network will be low, modest, high, or when sampling is intractable.

## 1 Introduction

A *Bayesian network*, *belief network*, or *directed acyclic graphical model*, is a probabilistic graph model [28] that has gained wide acceptance in several areas of artificial intelligence [31]. A Bayesian network represents a *joint probability distribution* (JPD) over a set of *statistical variables* and structurally models the *(conditional) independence relationships* between the variables as a *directed acyclic graph* (DAG). Efficient algorithms exist that perform *probabilistic inference* on Bayesian networks for reasoning with uncertainty and to solve decision problems with uncertain data.

Exact and approximate probabilistic Bayesian inference is known to be **NP**-hard [11, 12]. No single Bayesian inference algorithm or a class of algorithms is known to generally outperform others. Approximate inference algorithms are popular due to their *anytime property* [17] to produce an approximate result, possibly in real time [23]. For this reason, *stochastic sampling algorithms*, especially *importance sampling*, are among the most widely-used approximate inference methods [31]. Examples are SIS [32], AIS-BN [7], DIS [27], RIS [34] and EPIS-BN [35]. Importance sampling algorithms mainly differ in the choice of *importance function* to sample the JPD. All sampling algorithms perform poorly on Bayesian networks that are known to be hard for approximate inference. When zero probabilities are permitted in the JPDs, sampling algorithms become inefficient due to *sample rejection*; samples with zero probability do not contribute to the sum estimate and are therefore effectively rejected. Consequently, the sampling algorithm's performance is poor on Bayesian networks with *deterministic causalities*, i.e. networks that model JPDs with zero probabilities.

The *local variance bound* (LVB) [13] metric demarcates the boundary between the class of Bayesian networks with tractable approximations and those with intractable approximations. In certain cases the LVB of a Bayesian network can be used as a quantitative estimation of the expected rate of convergence of the approximate solution to the exact solution under stochastic sampling of the network.

However, the LVB metric is not applicable to measure the approximation hardness of Bayesian networks with deterministic causalities. The LVB metric is the ratio of the maximum to the minimum probability value of the *conditional probability table* (CPT) entries of the variables in the Bayesian network. Any zero entry in the CPT invalidates the LVB, i.e. when zero probabilities are permitted in the JPD. This means that the LVB is undefined for many real-world networks that model JPDs with zero probabilities, such as *Munin* [3], *Pathfinder* [24], *Andes* [10] and the *CPCS* [29] networks. Hence, their inference tractability classification fails and it is not possible to give an estimation of the hardness of approximate inference.

This paper presents the $k$-test, a metric to determine the approximation hardness of Bayesian inference. By contrast to the LVB metric, the $k$-test metric can be applied to Bayesian networks that model JPDs that include zero probabilities. This extends the approximation hardness estimation to an important class of real-world Bayesian networks. The $k$-test on a Bayesian network can be used to indicate when sample rejection rates are expected to be low, modest, or prohibitively high and sampling is intractable.

No polynomial-time algorithm exists that can filter samples $\mathbf{x}$ with $\Pr(\mathbf{x}) = 0$ from important sampling to prevent rejection. Exact determination of $\Pr(\mathbf{x}) = 0$ is known to be **NP**-complete. It turns out that approximate filtering of samples with $\Pr(\mathbf{x}) = 0$ is also **NP**-hard.

The remainder of this paper is organized as follows. Section 2 briefly introduces the Bayesian network formalism, inference by importance sampling, and the sample rejection problem. Section 3 introduces the $k$-test. Results of the $k$-test on real-world and benchmark Bayesian networks is presented in Section 4. Finally, Section 5 compares related work and summarizes our contribution.

## 2 Background

This section briefly introduces the Bayesian network formalism, approximate Bayesian inference by importance sampling, and the sample rejection problem.

### 2.1 Bayesian Networks

A Bayesian network is defined as follows.

**Definition 1** *A* Bayesian Network *$BN = (G, \Pr)$ consists of a directed acyclic graph (DAG) $G = (\mathbf{V}, \mathbf{A})$ with vertices $\mathbf{V}$, arcs $\mathbf{A} \subseteq \mathbf{V} \times \mathbf{V}$, and a JPD $\Pr$ over the discrete random variables $\mathbf{V}$ (represented by the vertices of $G$). $\Pr$ is defined by*

$$\Pr(\mathbf{V}) = \prod_{V \in \mathbf{V}} \Pr(V | \pi(V)) \quad,$$

*where $\pi(V)$ denotes the set of parents of a vertex $V$ and the* conditional probability tables (CPT) *of the BN assign domain-specific probabilities to $\Pr(V|\pi(V))$ for all $V \in \mathbf{V}$.*

In this paper, variables are indicated by uppercase letters and their states by lowercase letters. Sets (vertices, arcs, and states) are represented in boldface, e.g. $\mathbf{x} \in \mathit{Config}(\mathbf{X})$ denotes a state (configuration of values) $\mathbf{x}$ of a set of variables $\mathbf{X} \subseteq \mathbf{V}$. The set of *evidence variables* (or *vertices*) is denoted as $\mathbf{E}$, $\mathbf{E} \subset \mathbf{V}$ and the set $\mathbf{X} = \mathbf{V} \setminus \mathbf{E}$.

Exact probabilistic inference methods compute $\Pr(X|\mathbf{e})$ given *evidence* $\mathbf{e}$ for a set of evidence variables $\mathbf{E}$ directly from the network. Approximate inference methods estimate $\Pr(X|\mathbf{e})$.

Note that $\Pr(\mathbf{v}) = 0$ for configuration of variables $\mathbf{V}$ whenever the CPT entry $\Pr(v_i|\pi(v_i)) = 0$ for $V_i \in \mathbf{V}$. These zero entries are computationally problematic for Bayesian inference with importance sampling.

### 2.2 Approximate Bayesian Inference by Importance Sampling

Let $g(\mathbf{X})$ be a function over $m$ variables $\mathbf{X} = \{X_1, \ldots, X_m\}$ over some domain $\Omega \subseteq \mathbb{R}^m$, such that computing $g(\mathbf{X})$ for any $\mathbf{X}$ is feasible. Let $p$ be a probability density over $\mathbf{X}$. Consider the problem of estimating the integral

$$\mathbf{E}[g(\mathbf{X})|p] = \int_\Omega g(\mathbf{x}) p(\mathbf{x}) \, d\mathbf{x} \quad . \quad (1)$$

Assuming that $p$ is a density that is easy to sample from, the integral can be approximated by drawing a set of independent and identically distributed (i.i.d.) samples $\{\mathbf{x}_1, \ldots, \mathbf{x}_N\}$ from $p(\mathbf{X})$ and use these to compute the sample mean

$$\tilde{g}_N = \frac{1}{N} \sum_{i=1}^N g(\mathbf{x}_i) \quad . \quad (2)$$

According to the *strong law of large numbers*, $\tilde{g}_N$ almost surely converges to $\mathbf{E}[g(\mathbf{X})|p]$. The basic idea of importance sampling is to draw from a distribution other than $p$, say $I$, that is easier to sample from than $p$. We can rewrite (1) into

$$\mathbf{E}[g(\mathbf{X})|p] = \int_\Omega g(\mathbf{x}) \frac{p(\mathbf{x})}{I(\mathbf{x})} I(\mathbf{x}) \, d\mathbf{x} \quad , \quad (3)$$

where $I$ is often referred to as the *importance function*. The revised sample mean formula

$$\hat{g}_N = \sum_{i=1}^N [g(\mathbf{x}_i) w(\mathbf{x}_i)] \quad , \quad (4)$$

uses weights $w(\mathbf{x}_i) = \frac{p(\mathbf{x}_i)}{I(\mathbf{x}_i)}$. Again, $\hat{g}_N$ almost surely converges to $\mathbf{E}[g(\mathbf{X})|p]$.

**Assumption 1** *The distribution $I(\mathbf{X})$ is assumed to support $p(\mathbf{X})$ on $\Omega$. That is, $\forall \mathbf{x} \in \Omega : p(\mathbf{x}) > 0 \Rightarrow I(\mathbf{x}) > 0$. Furthermore, it is assumed that $\frac{0}{0} = 0$ in $w(\mathbf{x}_i) = \frac{p(\mathbf{x}_i)}{I(\mathbf{x}_i)}$.*

For more details refer to [30, 36].

The goal of the importance function $I$ is to approximate the posterior probability distribution $\Pr(X|\mathbf{e})$,

modeled by a network given some evidence **e** for **E**, without actually updating the network to the posterior, which is prohibitively expensive.

The importance function $I$ should be *tractable*. Here, "tractable" means that there exists a sampling order $\delta$, such that for any valid instantiation $x_{\delta(1)}, \cdots, x_{\delta(m)}$ of $X_{\delta(1)}, \cdots, X_{\delta(m)}$ $m \geq 1$, the complexity of computing $\Pr_I(x_{\delta(m+1)} \mid x_{\delta(1)}, \cdots, x_{\delta(m)})$ is polynomial. Here, $\Pr_I(\cdot)$ is the probability distribution induced by $I$. A sampling order $\delta$ that meets these requirements is called a *tractable sampling order for $I$*, or simply a *tractable order*.

Importance sampling methods generally require tractable sampling that is consistent with a Bayesian network's topological order of the vertices **V**. This is true for AIS-BN [7], EPIS-BN [35], and SIS [32]. By contrast, the tractable sampling order of DIS [27] is a reversed elimination order, which may or may not be consistent with a Bayesian network's topological order of the vertices.

A well-known problem of importance sampling on Bayesian networks with deterministic causalities is that the performance of sampling can be poor. When the JPD has zero probabilities, many samples may end up having zero weights $w(\mathbf{x}) = 0$ in the sampling process. These samples do not contribute to the sum estimate (4) and are effectively rejected. Such a sample is *inconsistent*, since **x** is an impossible event $w(\mathbf{x}_i) = 0 \Rightarrow p(\mathbf{x}_i) = 0$ by Assumption 1.

The *sample rejection problem* under evidential reasoning in a Bayesian network with evidence **e** is a judgement whether $\Pr(x_{\delta(1)} \cdots x_{\delta(m)}, \mathbf{e}) = 0$ for sample $x_{\delta(1)}, \cdots, x_{\delta(m)}$ and sampling order $\delta$.

## 3 The $k$-Test

In this section the $k$-test is presented to measure the approximation hardness of sampling Bayesian networks with deterministic causalities. The random $k$-SAT problem and the Satisfiability Threshold Conjecture are explained that form the basis for the $k$-test. An algorithm to efficiently compute the $k$-test ratio of a Bayesian network is given.

### 3.1 Hardness of Sample Rejection

Cooper [11] proved that computing $\Pr(\mathbf{x})$ for any **x** is **NP**-hard in general. Computing $\Pr(\mathbf{x})$ to classify inconsistent samples **x** with $\Pr(\mathbf{x}) = 0$ from consistent ones with $\Pr(\mathbf{x}) > 0$ is prohibitively expensive as a measure to determine the hardness of sampling.

Furthermore, the *approximate sample rejection problem* is too hard to be polynomial, as found by our proof in Appendix A. Dagum and Luby [12] proved that approximate Bayesian inference is **NP**-hard. However, their proof is not applicable to the sample rejection problem of Bayesian networks, because the JPDs of these networks are not strictly positive (an assumption used in their proof).

To empirically estimate the rejection rate requires a significant number of samples to be produced to cover the exponential state space of a network. Furthermore, the choice of a sampling algorithm may also influence the estimation of the ratio, such as by *CPT learning* adopted by state-of-the-art sampling algorithms.

Therefore, exact, approximate, and empirical determination of the hardness of sampling Bayesian networks exhibiting zero probabilities poses significant computational difficulties.

The sample rejection problem can be transformed into an equivalent random $k$-SAT problem, which forms the basis of our $k$-test.

### 3.2 Random $k$-SAT

Franco and Paull [15] first observed, among other important results, that random instances of the $k$-SAT problem undergo a "*phase transition*" as the ratio of clauses to variables passes through a threshold. Let $F_k^{n,m}$ denote a $k$-CNF with $n$ variables and $m$ $k$-clauses created by uniformly and randomly choosing $m$ clauses from the $C_k = 2^k \binom{n}{k}$ possible clauses. Franco and Paull [15] claim that $F_k^{n,m=rn}$ is *with high probability* (w.h.p. $\lim_{n\to\infty} \Pr(\epsilon_n) = 1$) unsatisfiable if $r \geq 2^k \ln 2$. The reasons are given as follows. Let **a** be a truth assignment and let $S_k = (2^k - 1)\binom{n}{k}$ be the number of $k$-clauses consistent with the given assignment **a**. Then for $F_k^{n,m}$, $\Pr(F_k^{n,m}(\mathbf{a}) = true) = \binom{S_k}{m}/\binom{C_k}{m} \leq (1 - 2^{-k})^m$ the expected number of satisfying truth assignments of $F_k^{n,m}$ is at most $2^n(1 - 2^{-k})^m = o(1)$ for $m/n \geq 2^k \ln 2$.

This result has led to the following popular conjecture.

*Satisfiability Threshold Conjecture.* For $k \geq 3$, there exists a constant $r_k$ such that

$$\lim_{n \to \infty} \Pr(F_k^{n,rn} \text{ is satisfiable}) = \begin{cases} 1, & \text{if } r < r_k, \\ 0, & \text{if } r > r_k \end{cases} \quad (5)$$

Since 1990 much work has been done on this conjecture to narrow the threshold $r_k$. One class of methods is based on mathematical analysis. In the milestone paper [16], Friedgut used the second moment method to prove the existence of a nonuniform satisfiability threshold, i.e. a sequence $r_k(n)$, around which the probability of satisfiability goes from 1 to 0. Inspired by [16], Dimitris and Cristopher [1] further narrowed the threshold around $O(2^{k-1} \ln 2)$.

Another method is to design and analyze a polynomial algorithm that can find a truth assignment *with uniformly positive probability* (w.u.p.p. $\lim_{n\to\infty} \inf \Pr(\epsilon_n) > 0$) or w.h.p., if, for a satisfiable $F_k^{n,rn}$, $r$ is smaller than the lower bound of $r_k$. In [6] and [8] this method is used to narrow the lower bound of $r_k$ to $O(2^k/k)$. The current best result is from [9], which not only presents a polynomial algorithm that finds a satisfying truth assignment w.h.p., if $r < (1 - \epsilon_k)2^k \ln(k)/k$, where $\epsilon_k \to 0$ and $k > 10$, but also points out that if $r$ is above $O(2^k \ln(k)/k)$ no polynomial algorithm is known to find a satisfying truth assignment with probability $\Omega(1)$ – neither on the basis of rigorous or empirical analysis, or any other evidence.

### 3.3 The $k$-Test Algorithm

If a one-to-one mapping from the set of consistent samples from a Bayesian network to a set of satisfying truth assignments of a satisfiable $k$-CNF $F_k^{n,m}$ can be constructed, then the clause density $r = m/n$ can be compared to the threshold value $2^k/k$ to estimate the hardness of the rejection problem to sample the network. The $k$-test ratio of a Bayesian network is the ratio of the clause density $r$ to the threshold value $2^k/k$. What remains is to devise an efficient construction method for the $k$-CNF of a Bayesian network to determine the $k$-test ratio $r : 2^k/k$. The construction of a $k$-CNF $F_k^{n,m}$ should satisfy the following requirements:

1. The satisfying truth assignments of the $k$-CNF $F_k^{n,m}$ model all and only the consistent configurations of the Bayesian network.

2. The $k$-CNF $F_k^{n,m}$ should be minimal. That is, there should not be too many unnecessary binary variables introduced in the $k$-CNF or too many unnecessary clauses added to the $k$-CNF.

3. The construction of the $k$-CNF and calculation of the clause density $r$ should be performed in polynomial time.

In the following, we define an efficient conversion from a Bayesian network to a $k$-CNF $F_k^{n,m}$ that satisfies these three requirements.

Let $BN = (G, \Pr), G = (\mathbf{V}, \mathbf{A})$ be a Bayesian network and $\mathbf{E}$ the evidence, where $\mathbf{e}$ is the evidence configuration. Let $\|V_i\|$ denote the number of states of $V_i \in \mathbf{V}$. The conversion of the $BN$ to a Boolean formula proceeds in two steps.

*Step 1.* Convert all variables $\mathbf{V}$ to Boolean variables by log encoding [18]. For each $V_i \in \mathbf{V}$, we create a set of Boolean variables:

$$\{X_{i_1}, X_{i_2}, \cdots, X_{i_{\lceil \log_2 \|V_i\| \rceil}}\} \quad (6)$$

Map each $X_{i_k}$ to the $k^{\text{th}}$ bit of the binary representation of $V_i$'s discrete value. For example, if $\|V_i\| = 5$, then $V_i = 3$ is mapped to $(X_{i_3} = 0, X_{i_2} = 1, X_{i_1} = 1)$.

*Step 2.* Construct the $k$-CNF formula $F_k^{n,m}$. There are three different types of clauses in the resulting $k$-CNF: the *CPT Clauses*, the *Variable Clauses*, and the *Evidence Clauses*. The aim is to construct a $k$-CNF that is minimal, which is accomplished as follows.

*CPT Clauses:* each zero entry in a CPT of the Bayesian network represents a constraint that is translated into a disjunctive clause. For example, assume $\|V_i\| = 5, \|V_j\| = 3$, then we can translate $\Pr(V_i = 3 \mid V_j = 2) = 0$ to $\neg(\neg X_{i_3} \wedge X_{i_2} \wedge X_{i_1} \wedge X_{j_2} \wedge \neg X_{j_1}) = X_{i_3} \vee \neg X_{i_2} \vee \neg X_{i_1} \vee \neg X_{j_2} \vee X_{j_1}$.

*Variable Clauses:* for each $V_i \in \mathbf{V}$, if $\|V_i\|$ is not a power of 2, then certain assignments of $V_i$'s translated Boolean variables do not satisfy the formula. For example, if $\|V_i\| = 5$, then $\{X_{i_1} = 1, X_{i_2} = 1, X_{i_3} = 1\}$ is not a valid assignment, since 7 is not a valid value of $V_i$. We add a clause for each invalid assignment. For the previous example, this means that $\neg(X_{i_1} \wedge X_{i_2} \wedge X_{i_3})$ is added. However, if this mapping is done naïvely, the result may yield an exponentially large set of clauses, which clearly does not lead to a minimal $k$-CNF. Consider $\|V_j\| = 2^m + 1$, then for $V_j$ as many as $2^m - 1$ clauses will be added to the $k$-CNF formula. In the worst case, this leads to a $(2^m - 1)/(m + 1)$ clause/variable ratio. Fortunately, we can take advantage of the fact that all valid discrete values of $V_j$ are smaller than $\|V_j\|$ to minimize the number of clauses. Let $1b_m \cdots b_1$ be the binary representation of $\|V_j\| - 1$. Then, for $k = m, \ldots, 1$ such that $b_k = 0$, we have that

$$X_{j_{m+1}} = 1 \wedge \cdots \wedge X_{j_{k+1}} = b_{k+1} \to X_{j_k} = 0. \quad (7)$$

Formula (7) can be converted to a disjunctive clause by $A \to B \Leftrightarrow \neg A \vee B$, since $X_l = 1(0)$ is simply equivalent to $X_l = true(false)$. In the worst case, the variable clauses only contribute $m/(m+1)$ clauses to the clause density.

*Evidence Clauses:* for each $V_i \in \mathbf{E}$ with state $V_i = e_i$, let $b_m b_{m-1} \cdots b_1$, $m = \|V_i\|$, be the binary representation of $e_i$. Add the clause $X_{i_m} = b_m \wedge X_{i_{m-1}} = b_{m-1} \wedge \cdots \wedge X_{i_1} = b_1$ to the $k$-CNF.

To compute the $k$-test ratio $r : 2^k/k$ of a Bayesian network, only the density $r$ of the $k$-CNF clauses is

**Procedure** k-Test $(BN, \mathbf{E})$
**Input**: $BN = (G, CPT)$, evidence set $\mathbf{E} \subseteq \mathbf{V}(G)$
**Output**: $k$, clause density $r$
**begin**
$\quad n \leftarrow \sum_{V_i \in \mathbf{V}(G)} \lceil \log_2 \|V_i\| \rceil$
$\quad m \leftarrow |\mathbf{E}|$
$\quad k \leftarrow \max_{E \in \mathbf{E}} \lceil \log_2 \|E\| \rceil$
$\quad$**foreach** $V_i \in \mathbf{V}(G)$ **do**
$\quad\quad k_i \leftarrow \lceil \log_2 \|V_i\| \rceil$
$\quad\quad b \leftarrow \|V_i\| - 1$
$\quad\quad$**while** $b > 0$ **do**
$\quad\quad\quad$**if** $b \mod 2 = 0$ **then**
$\quad\quad\quad\quad m \leftarrow m + 1$
$\quad\quad\quad\quad$**if** $k_i > k$ **then** $k \leftarrow k_i$
$\quad\quad\quad$**end**
$\quad\quad\quad b \leftarrow \lfloor b/2 \rfloor$
$\quad\quad\quad k_i \leftarrow k_i - 1$
$\quad\quad$**end**
$\quad\quad k_{\max} \leftarrow \sum_{V_j \in \pi[V_i]} \lceil \log_2 \|V_j\| \rceil + \lceil \log_2 \|V_i\| \rceil$
$\quad\quad$**foreach** $\Pr_j \in CPT[V_i]$ **do**
$\quad\quad\quad$**if** $\Pr_j = 0$ **then**
$\quad\quad\quad\quad m \leftarrow m + 1$
$\quad\quad\quad\quad$**if** $k_{\max} > k$ **then** $k \leftarrow k_{\max}$
$\quad\quad\quad$**end**
$\quad\quad$**end**
$\quad$**end**
$\quad r \leftarrow m/n$
**end**

**Algorithm 1**: $k$-Test

| Bayesian Network | Nodes $|\mathbf{V}|$ | Arcs $|\mathbf{A}|$ | Ratio $r : 2^k/k$ | Rejection Rate % |
|---|---|---|---|---|
| cpcs360b | 360 | 729 | 0.000 | 0% |
| cpcs422b | 422 | 867 | 0.001 | 0% |
| cpcs179 | 179 | 239 | 0.006 | 0% |
| Andes | 223 | 338 | 0.018 | 27% |
| Munin | 1,041 | 1,397 | 0.112 | 97% |
| Pathfinder | 109 | 195 | 0.173 | 98% |

Table 1: Results for Real-World Bayesian Networks

| Bayesian Network | Nodes $|\mathbf{V}|$ | Arcs $|\mathbf{A}|$ | Ratio $r : 2^k/k$ | Rejection Rate % |
|---|---|---|---|---|
| BN_102uai | 76 | 210 | 0.000 | 0% |
| BN_28uai | 24 | 30 | 0.000 | 0% |
| BN_88uai | 422 | 867 | 0.001 | 0% |
| BN_90uai | 422 | 867 | 0.001 | 0% |
| BN_92uai | 422 | 867 | 0.001 | 0% |
| BN_18uai | 2,127 | 3,595 | 0.002 | 7.2% |
| BN_16uai | 2,127 | 3,595 | 0.002 | 7.8% |
| BN_70uai | 2,315 | 4,318 | 0.262 | 100% |
| BN_71uai | 1,740 | 3,012 | 0.270 | 100% |
| BN_72uai | 2,155 | 3,686 | 0.275 | 100% |
| BN_75uai | 1,820 | 3,328 | 0.280 | 100% |
| BN_73uai | 2,140 | 3,538 | 0.300 | 100% |
| BN_74uai | 749 | 1,214 | 0.307 | 100% |
| BN_76uai | 2,155 | 3,686 | 0.320 | 100% |
| BN_69uai | 777 | 1,322 | 0.460 | 100% |

Table 2: Results for Benchmark Bayesian Networks

required. Hence, the $k$-test algorithm is a simplified version of the $k$-CNF construction algorithm, in which the number of variables $n$, number of $k$-clauses $m$, and $k$ are computed directly from a Bayesian network as shown in Algorithm 1. The algorithm determines $k$ and the clause density $r = m/n$ given a Bayesian network $BN$ and set of evidence variables $\mathbf{E}$. The time complexity of the algorithm is linear to the sum of the sizes of the CPTs in the network. From our empirical results reported in the next section, we found that the $k$-test only takes seconds to compute for large networks with hundreds of variables.

## 4 Results

In this section the $k$-test is experimentally verified. Two classes of Bayesian networks were used in the experiments: "real-world" networks and benchmark networks. The real-world networks are shown in Table 1. *Munin* [3], *Pathfinder* [24], *Andes* [10] and *CPCS* [29] are networks with significant levels of determinism, suggesting difficulties with high rejection rates to sample them. A selection[1] of synthetic benchmark networks from the UAI contest [5] is shown in Table 2. Both tables show the number of variables and arcs of the network, the directly computed $k$-test ratio $r : 2^k/k$, and the average rejection rate of naïve sampling. To eliminate any bias of advanced sampling techniques toward any of these networks, naïve importance sampling (*likelihood sampling*) is used in this study. In this way, rejection rates purely depend on the properties of the JPD of the Bayesian network, not on adaptive CPT learning-based optimizations to sample the network as is performed by SIS, AIS-BN, and other advanced algorithms. Furthermore, for each Bayesian network, we randomly generated 50 test cases, and for each test case a random set of 10 to 20 evidence variables and instantiations are randomly selected. The number of samples is 60,000 for each test case.

Figure 1 compares the directly computed ratio $r : 2^k/k$ to the empirically-established rejection rate of likelihood importance sampling for all Bayesian Networks used in this study. In the figure, a rejection rate of 1.0 means that all samples are rejected in the sampling process (100%).

From these results it can be concluded that the $k$-test ratio $r : 2^k/k$ accurately predicts when the sampling rejection rate will be modest, high, or reaches 100% and sampling becomes intractable. When the $k$-test ratio $< 0.1$, sample rejection rates are modest or zero, see Tables 1 and 2. The sampling efficiency is poor when the $k$-test ratio reaches 0.1. When the $k$-test ratio $> 0.2$, sampling is intractable. No importance sampling algorithm can generate consistent samples for the synthetic BNs from BN_69uai to BN_76uai in reasonable time. Likelihood sampling does not even yield a single valid sample in thousands of samples, see Table 2.

---
[1] We selected networks that are easy to hard to sample.

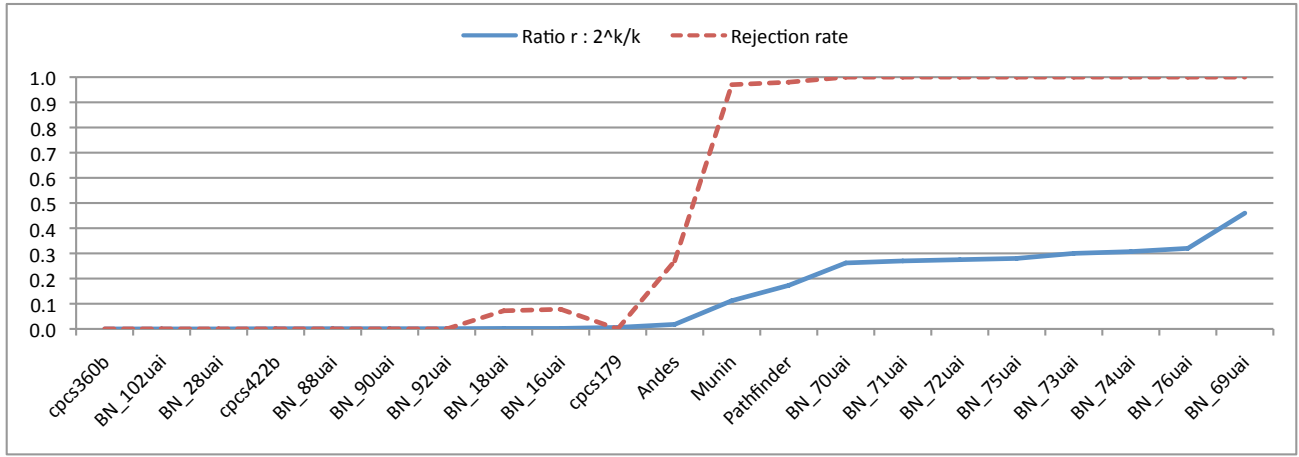

Figure 1: Ratio $r : 2^k/k$ and Rejection Rate for all Bayesian Networks in this Study

When the $k$-test ratio $r : 2^k/k$ of a network is between 0.005 and 0.200, improved importance sampling methods can be used to attempt to lower the sampling rejection rate and thus improve efficiency of sampling. Example sampling improvements are SIS [32], AIS-BN [7], DIS [27], RIS [34] and EPIS-BN [35].

State-of-art importance sampling algorithms are known to perform well on *Andes* ($k$-test ratio 0.018, the average rejection ratio of AIS-BN is 13.9%) and *Pathfinder* ($k$-test ratio 0.173, the average rejection ratio of AIS-BN is 49.5%) as these algorithms mitigate the rejection problem. However, their performance is mixed on *Munin* ($k$-test ratio 0.112, the average rejection ratio of AIS-BN is 96.1% and the average rejection ratio of EPIS-BN is 28.5%).

Easy-to-sample networks do not require sophisticated sampling techniques. For those networks the $k$-test ratio $< 0.005$. Indeed, sampling the *CPCS* networks (Table 1) incurs no rejection overhead. Simple sampling methods suffice for these networks.

## 5   Conclusions

This paper introduced the $k$-test to measure the hardness of stochastically sampling Bayesian networks that exhibit zero probabilities. Such networks have deterministic causalities defined by the zeros in the conditional probability tables (CPT), which results in samples being rejected. To empirically estimate the rejection rate requires a significant number of samples to be produced to cover the exponential state space of a network. It also requires the use of a set of sampling algorithms to eliminate algorithm bias (such as CPT learning effects). By contrast, the $k$-test is a linear-time algorithm to determine the hardness of stochastically sampling a Bayesian network and is a good estimator of the rejection rate for any sampling algorithm. The metric identifies networks for which rejection rates will be low, modest, high, or when sampling is intractable. The $k$-test is based on recent advances in random $k$-SAT analysis. Experimental results for real-world and benchmark networks show the experimental validity of the $k$-test.

Sampling algorithms have been modified and improved by many authors to mitigate the generation of inconsistent samples and limit the overhead of sample rejection. The rejection problem in importance sampling has been extensively studied in the work on *adaptive sampling schemes* [7], in the context of *constraint propagation* [20], and *Boolean satisfiability problems* [21]. A restricted form of constraint propagation can be used to reduce the amount of rejection [19]. An approach to circumvent the rejection problem by systematically searching for a nonzero weight sample for constraint-based systems was introduced in [20]. The proposed backtracking algorithm, *SampleSearch* was further improved in [21] and shown to generate a *backtrack-free* distribution. In [22], the *SampleSearch* method is fur-ther generalized as a scheme in the framework of *mixed networks* [14, 26]. However, the exact rejection problem is **NP**-complete and the approximate rejection problem is too hard to be polynomial as we proved in this paper. Because Bayesian networks are special cases of mixed networks, we believe that Corollary 1, Theorem 2 and Theorem 1 can be generalized to mixed networks.

Although most state-of-art importance sampling algorithms have a capability to reduce the generation of inconsistent samples, in worst case they still fail to generate sufficient useful samples in reasonable time. It is therefore critical to identify Bayesian networks that are hard to sample, e.g. by using the $k$-test.

The LVB [13] metric demarcates the boundary between the class of Bayesian networks with tractable approximations and those with intractable approximations. LVB is undefined for many real-world networks that model JPDs with zero probabilities, such as *Munin* [3], *Pathfinder* [24], *Andes* [10] and the *CPCS* [29] networks. The $k$-test compliments the LVB by measuring the approximation hardness of these and many other Bayesian networks with deterministic causalities, i.e. networks that model JPDs with zero probabilities. LVB measures the hardness of sampling caused by strictly-positive extreme probability distribution, whereas the $k$-test measures the difficulties of sampling induced by the rejection problem. Currently there is no satisfactory combination of these two measurements that provides a general metric to measure the hardness of sampling a Bayesian network. This will be an interesting and challenging forthcoming work, because a combined metric enables the measurement of the sampling hardness of networks that exhibit both zero and close-to-zero probabilities.

## A  Appendix

### A.1  Probabilistic Turing Machines

The *Probabilistic Turing Machine* (PTM) formulation is used in the complexity analysis of approximate algorithms and probabilistic algorithms. We briefly introduce PTM and class **RP**, see [4] for more details.

**Definition 2** *The Probabilistic Turing Machine (PTM) is a Turing machine with two transition function sets $\lambda_0, \lambda_1$. To execute a PTM M on an input x, we choose in each step with probability $\frac{1}{2}$ to apply the transition function in $\lambda_0$ and with probability $\frac{1}{2}$ to apply $\lambda_1$. This choice is made independently.*

*The machine M only outputs 1 ("accepted") or 0 ("rejected"). We denote by $M(x)$ the random variable corresponding to the value M outputs at the end of execution. For a function $T : \mathbb{N} \to \mathbb{N}$, we say that M runs in $T(n)$-time if for any input x, M halts on x within $T(|x|)$ steps regardless of the random choices it makes.*

**Definition 3** $\mathbf{RTIME}(T(n))$ *contains every language L for which there is a PTM M running in $T(n)$ time such that*

$$\begin{array}{l} x \in L \Rightarrow \Pr(M(x) = 1) \geq \frac{2}{3} \\ x \notin L \Rightarrow \Pr(M(x) = 0) = 1 \end{array} \quad (8)$$

*We define* $\mathbf{RP} = \bigcup_{c>0} \mathbf{RTIME}(n^c)$

Obviously $\mathbf{RP} \subseteq \mathbf{NP}$.

### A.2 Hardness of the Sample Rejection Problem

First, we give a formal definition of sample rejection problem:

**Definition 4** *For any tractable[2] importance function I with tractable sampling order $\delta$, For a BN, $\mathbf{e}$ are evidences and $\mathbf{e} \neq \varnothing$. I is an importance function and $\delta$ is a sampling order. $\Pr_\mathbf{e}(\cdot)$ is the BN's posterior probability distribution ($\Pr_\mathbf{e}(\cdot) \equiv 0$, if $\Pr(\mathbf{e}) = 0$). the Rejection Problem of I with $\delta$ is defined as: let $\mathbf{e} \neq \varnothing$, $\Pr(\mathbf{e}) > 0$, be the observed evidence[3], if $\Pr_I(x_{\delta(1)}, \cdots, x_{\delta(m-1)}) > 0$ ($m \geq 1$), then $\forall x \in Config(X_{\delta(m)})$ determine whether $\Pr_\mathbf{e}(x_{\delta(1)}, \cdots, x_{\delta(m-1)}, x) > 0$.*

Def. 4 is reasonable, because if during sampling of the $m^{\text{th}}$ variable $X_{\delta(m)}$ the rejection problem is solved, then we can pick up a $x_{\delta(m)}$ from $Config(X_{\delta(m)})$ such that $\Pr_\mathbf{e}(x_{\delta(1)}, \cdots, x_{\delta(m)}) > 0$. This process repeats until we find a consistent sample.

Note that Def. 4 requires nothing when $\Pr(\mathbf{e}) = 0$. Thus, the exact rejection algorithm is a partial function (*Rejection Function*) $\Gamma_\delta : \Omega_\delta \to \{0,1\}$, $\Omega_\delta = Config(\mathbf{E}) \times \bigcup_{i=1}^{|\mathbf{X}|} Config(X_{\delta(1)} \cdots X_{\delta(i)})$. If $\Pr(\mathbf{e}) = 0$,

---
[2]Only tractable importance functions with tractable sampling orders are considered, because the sampling process should be polynomial.

[3]$\mathbf{e} \neq \varnothing$ because generating a consistent sample on a BN without evidence is a trivial problem.

$\Gamma_\delta(\mathbf{e}, \cdot)$ is undefined (could be any); if $\Pr(\mathbf{e}) > 0$,

$$\Gamma_\delta(\mathbf{e}, x_{\delta(1)} \cdots x_{\delta(m)}) = \left\{ \begin{array}{ll} 1, & \Pr_\mathbf{e}(x_{\delta(1)} \cdots x_{\delta(m)}) > 0 \\ 0, & \Pr_\mathbf{e}(x_{\delta(1)} \cdots x_{\delta(m)}) = 0 \end{array} \right.$$

Here, the sampling probability distribution $\Pr_I^\Gamma$, induced by importance function $I$ and rejection function $\Gamma_\delta$, is obtained by:

$$\begin{array}{l}\Pr_I^\Gamma(x_{\delta(m)} \mid x_{\delta(1)} \cdots x_{\delta(m-1)}) = \\ \frac{\Pr_I(x_{\delta(m)}|x_{\delta(1)}\cdots x_{\delta(m-1)}) \times \Gamma_\delta(x_{\delta(m)}|x_{\delta(1)}\cdots x_{\delta(m-1)})}{\sum_{x \in Config(X_{\delta(m)})} \Pr_I(x|x_{\delta(1)}\cdots x_{\delta(m-1)}) \times \Gamma_\delta(x_{\delta(1)}\cdots x_{\delta(m-1)},x)}.\end{array} \quad (9)$$

Assuming $\Pr(\mathbf{e}) > 0$, a rejection algorithm solves the rejection problem in computable $T(n)$ time, for input with length $n$. The ill-defined case $\Pr(\mathbf{e}) = 0$ can be bounded by counting down from $T(n)$ and returning a random decision when the counter reaches 0. Hence, it can be assumed that rejection algorithms are time bounded.

The sample rejection problem is **NP**-complete. That is, there is no polynomial time algorithm that generally classifies consistent and inconsistent samples from Bayesian networks.

**Lemma 1** *For any tractable importance function I with tractable sampling order $\delta$, the rejection problem is in* **NP**.

Lemma 1 is straightforward because verifying whether a sample is consistent is $\mathcal{O}(n)$. To prove that the rejection problem is **NP**-complete, we reduce the 3SAT problem into the rejection problem and the reduction is polynomial.

**Corollary 1** *For any tractable importance function I with tractable sampling order $\delta$, the sample rejection problem is* **NP**-*complete.*

**Proof.** This follows from [11]. For any 3CNF $\mathcal{F}$, we convert $\mathcal{F}$ to PIBNET by [11]. Assume that $Y$ (the descendent vertex of all the other vertices in PIBNET, which represents the value of $\mathcal{F}$) is the only evidence and that $Y = True$. For any tractable importance function $I$ with sampling order $\delta$, if the rejection problem can be resolved in polynomial time, we can obtain a full sample by Eq. (9) in polynomial time. In case the denominator of Eq. (9) is zero, we can randomly pick a value of $X_{\delta(m)}$. If $\Pr(Y = True) > 0$, then $\Gamma$ is well defined and ensures one consistent sample (its weight $> 0$). If $\Pr(Y = True) = 0$ the generated sample must be inconsistent. Hence, we can differentiate $\Pr(Y = True) = 0$ from $\Pr(Y = True) > 0$ by solving the rejection problem. From Lemma 1 the rejection problem is in **NP**, and therefore the rejection problem is **NP**-complete. □

Approximate sample rejection problem is also **NP**-hard. A rejection algorithm is called *approximate*, if it may accept $x_{\delta(m)}$, even when $\Pr(\mathbf{e}, x_{\delta(1)} \cdots x_{\delta(m)}) = 0$, for $\Pr(\mathbf{e}) > 0 \land \Pr_I(x_{\delta(1)} \cdots x_{\delta(m-1)}) > 0$. Still, if $\Pr(\mathbf{e}, x_{\delta(1)} \cdots x_{\delta(m)}) > 0$ then an approximate algorithm must accept it to avoid biased sampling. In other words, an approximate rejection algorithm is an one-side error approximation. Furthermore, an *approximate rejection function* $\hat{\Gamma}_\delta^A$ of an approximate rejection algorithm $A$ is defined like $\Gamma_\delta$: if $\Pr(\mathbf{e}) > 0$,

$$\hat{\Gamma}_\delta^A(\mathbf{e}, x_{\delta(1)} \cdots x_{\delta(m)}) = \begin{cases} 1, & \text{if } A \text{ accepts } x_{\delta(1)} \cdots x_{\delta(m)} \\ 0, & \text{else} \end{cases}$$

When there is no confusion, $\hat{\Gamma}_\delta^A$ is simplified in this text as $\hat{\Gamma}_\delta$ or simply $\hat{\Gamma}$.

Let $\Pr_I^{\hat{\Gamma}}$ denote the sampling probability distribution induced by importance function $I$ and $\hat{\Gamma}$ ($\Pr_I^{\hat{\Gamma}}$ can be computed from Eq. 9 where $\Gamma_\delta$ is replaced by $\hat{\Gamma}$). Then, for inconsistent sample set $\Omega_{\mathbf{e}} = \{\mathbf{x} \mid \mathbf{x} \in Config(\mathbf{X}) \land \Pr_{\mathbf{e}}(\mathbf{x}) = 0\}$, $\Pr_I^{\hat{\Gamma}}(\Omega_{\mathbf{e}}) > 0$. $\Pr_I^{\hat{\Gamma}}(\Omega_{\mathbf{e}})$ gives the probability that inconsistent samples are generated or the average percentage of inconsistent samples over all the samples. If $\Pr_I^{\hat{\Gamma}}(\Omega_{\mathbf{e}}) = 0$, we get an exact algorithm. Clearly for an approximate rejection algorithm, the smaller $\Pr_I^{\hat{\Gamma}}(\Omega_{\mathbf{e}})$ is, the better is the approximation. We hope that there exists a polynomial $\xi$ *up-bounded* ($\xi$ *up-bounded* means $\exists \xi$, $0 < \xi < 1$, for any BN and any possible evidences $\mathbf{e}$ ($\Pr(\mathbf{e}) > 0$), $\Pr_I^{\hat{\Gamma}}(\Omega_{\mathbf{e}}) < \xi$) approximate rejection algorithm, so that $n$ consistent samples can be retrieved from $\frac{n}{\xi}$ samples with high probability.

Furthermore, since both the algorithm for computing the importance function and the rejection algorithm may be stochastic, for a BN and evidence $\mathbf{e}$, the $\Pr_I^{\hat{\Gamma}}(\Omega_{\mathbf{e}})$ may be a random variable. Thus, it is reasonable to define *randomly up-bounded* or $(\xi, \sigma)$ *up-bounded* as $\exists \xi \sigma$, $0 < \xi < 1 \land 0 \leq \sigma < 1$, for any BN and BN's evidence $\mathbf{e}$ ($\Pr(\mathbf{e}) > \mathbf{0}$), such that $\Pr[\Pr_I^{\hat{\Gamma}}(\Omega_{\mathbf{e}}) < \xi] \geq 1 - \sigma$. In other words, we relax the requirement of $\xi$ *up-bounded* to the case where $\xi$ *up-bounded* is satisfied with high probability. However Theorem 1 gives a pessimistic answer.

**Theorem 1** *If there exists a polynomial $(\xi, \sigma)$ up-bounded approximate rejection algorithm for some tractable importance function $I$ with tractable sampling order $\delta$, then $\mathbf{NP} \subseteq \mathbf{RP}$.*

**Proof.** Assume that $\hat{\Gamma}_\delta^A$ is the approximate rejection function of approximate rejection algorithm $A$ and $\Pr_I^{\hat{\Gamma}}$ is the sampling probability distribution induced by importance function $I$ and $\hat{\Gamma}_\delta^A$. Let $\xi$, $0 < \xi < 1$ and $\sigma$, $0 \leq \sigma < 1$. Then for any 3CNF $\mathcal{F}$, we convert $\mathcal{F}$ to PIBNET by the method of [11]. In PIBNET, value of vertex $Y$ is corresponding to value of $\mathcal{F}$. For $Y = True$ as the only evidence. Then we independently execute importance sampling process $m$ ($m > -\frac{\ln 3}{\ln(\sigma + (1-\sigma)\xi)}$) times with tractable importance function $I$ and approximate rejection algorithm $A$. Since both generating importance function $I$ and rejection algorithm $A$ maybe stochastic, we may obtain $m$ different $\Pr_I^{\hat{\Gamma}}(\cdot)$. Then we generate a sample from each $\Pr_I^{\hat{\Gamma}}(\cdot)$. If any one of the $m$ samples is consistent, we accept $\mathcal{F}$. If none of them is consistent, we reject.

- If $\mathcal{F}$ is unsatisfiable, no consistent sample can be generated.
- If $\mathcal{F}$ is satisfiable, $\Pr[\Pr_I^{\hat{\Gamma}}(\Omega_{\mathbf{e}}) < \xi] \geq 1 - \sigma \Rightarrow \Pr(\mathcal{F} \ rejected) < \sum_{i=0}^{m} \binom{m}{i} \sigma^{m-i}(1-\sigma)^i \xi^i$. Since $\sum_{i=0}^{m} \binom{m}{i} \sigma^{m-i}(1-\sigma)^i \xi^i = (\sigma + (1-\sigma)\xi)^m$, $\sigma + (1-\sigma)\xi < 1$ and $m > -\frac{\ln 3}{\ln(\sigma + (1-\sigma)\xi)}$, so $\Pr(\mathcal{F} \ rejected) < \frac{1}{3}$.

Hence, $\mathbf{NP} \subseteq \mathbf{RP}$, if a polynomial $(\xi, \sigma)$ up-bounded approximate rejection algorithm exists. $\square$

Since $\mathbf{RP} \subseteq \mathbf{P}_{/\mathbf{poly}}$ [2] and $\mathbf{NP} \subseteq \mathbf{P}_{/\mathbf{poly}} \Rightarrow \mathbf{PH} = \Sigma_2$ [33, 25], if a polynomial $(\xi, \sigma)$ up-bounded approximate rejection algorithm exists, then $\mathbf{PH}$ will collapse to $\Sigma_2$. It is widely believed that $\mathbf{PH}$ does not collapse to $\Sigma_2$, thus a polynomial $(\xi, \sigma)$ up-bounded approximate algorithm is unlikely to exist. Furthermore Theorem 1 tells that all tractable importance sampling algorithms may fail to generate sufficient samples in polynomial time for certain cases. Another implication of Theorem 2 is that for any tractable importance function $I$ sampling order $\delta$, no polynomial approximate rejection algorithm satisfies $\exists \xi \sigma : 0 < \xi < 1 \land 0 \leq \sigma < 1$, for any BN and valid $\mathbf{e}$ such that $\Pr_I^{\hat{\Gamma}}(\Omega_{\mathbf{e}}) < \xi \Pr_I(\Omega_{\mathbf{e}})$ with probability larger than $1 - \sigma$ unless $\mathbf{PH}$ collapses to $\Sigma_2$. Because $\Pr_I(\Omega_{\mathbf{e}}) < 1$, so $\Pr_I^{\hat{\Gamma}}(\Omega_{\mathbf{e}}) < \xi \Pr_I(\Omega_{\mathbf{e}}) \Rightarrow \Pr_I^{\hat{\Gamma}}(\Omega_{\mathbf{e}}) < \xi$. In other words, no polynomial approximate rejection algorithm can help importance sampling to reduce inconsistent samples with high probability.

Because $(\xi, 0)$ up-bounded is equivalent to $\xi$ up-bounded, It is straightforward to get the corollary 2.

**Corollary 2** *If there exists a polynomial $\xi$ up-bounded approximate rejection algorithm for some tractable importance function $I$ with tractable sampling order $\delta$, then $\mathbf{NP} \subseteq \mathbf{RP}$.*